# Activity Modeling in Smart Home using High Utility Pattern Mining over Data Streams

[1] Menaka Gandhi . J, [2] Gayathri K . S

[1] Computer Science & Engineering, Anna University, Sri Venkateswara College of Engineering
Chennai, Tamilnadu 602105, India

[2] Computer Science & Engineering, Anna University, Sri Venkateswara College of Engineering
Chennai, Tamilnadu 602105, India

**Abstract**

Smart home technology is a better choice for the people to care about security, comfort and power saving as well. It is required to develop technologies that recognize the Activities of Daily Living (ADLs) of the residents at home and detect the abnormal behavior in the individual's patterns. Data mining techniques such as Frequent pattern mining (FPM), High Utility Pattern (HUP) Mining were used to find those activity patterns from the collected sensor data. But applying the above technique for Activity Recognition from the temporal sensor data stream is highly complex and challenging task. So, a new approach is proposed for activity recognition from sensor data stream which is achieved by constructing Frequent Pattern Stream tree (FPS - tree).  FPS is a sliding window based approach to discover the recent activity patterns over time from data streams. The proposed work aims at identifying the frequent pattern of the user from the sensor data streams which are later modeled for activity recognition. The proposed FPM algorithm uses a data structure called Linked Sensor Data Stream (LSDS) for storing the sensor data stream information which increases the efficiency of frequent pattern mining algorithm through both space and time. The experimental results show the efficiency of the proposed algorithm and this FPM is further extended for applying for power efficiency using HUP to detect the high usage of power consumption of residents at smart home.
.

*Keywords: Data Mining, Interactive Mining, Linked Sensor data stream, Activity Recognition, Frequent patterns, High Utility Patterns.*

## 1. Introduction

Ambient assisted living is an emerging area which focuses on helping elderly people to function independently at home.  Smart home is the combination of both technology and services through the process of networking which is set up at home for a better quality of living [1]. The need for such technology is due to the aging of the population, high quality of living, costly formal health care and the importance of the residents that the others at their own homes place.        Smart home helps the residents in improving home comfort, convenience, security and energy management [2] [11].

A smart home appears "intelligent" because its computer systems can monitor so many aspects of daily living. Residents have to complete their Activities of Daily Living (ADLs) [15] such as eating, dressing, sleeping, cooking etc. The monitoring of activities enables to detect the undesired situations that the residents face which can be used to trigger an emergency mechanism [16]. So, it is required to discover the users common behavior and predict his / her future actions in smart home. Therefore, activity records can be effectively analyzed to determine the behavior patterns [3].

Activity Modeling [17] plays an important role in Activity Recognition which is a complex task without human supervision. The main reason is that humans do not perform the activities in the same sequence as they did before. i.e There will be uncertainty in the human behavior. To solve such problem, AI (Artificial Intelligence) techniques are widely used. The main goal of this work is to observe the actions and activities that the residents perform and  model those activities and discover interesting patterns of activities [1]. This method of Activity Modeling can help the residents in reduced power consumption of electronic components [7].

## 2. Related Work

Mining the activity sequences or patterns is an important task in the smart environment. There are two types of pattern mining such as Frequent Pattern Mining and High Utility Pattern Mining [4]. It becomes possible to predict the behavior of the residents and also perform Activity modeling by discovering frequent patterns, temporal sensor data stream [5] and the expected utilities. There were various algorithms proposed to find those frequent patterns. Apriori was the basic algorithm for finding frequent patterns of activities proposed by R.Agrawal and R.Srikant in 1994 [8] [10]. This algorithm uses prior knowledge to find out the frequent patterns. The downward closure property is used to prune the infrequent patterns. The property states that if a pattern is infrequent, then all of its super patterns must be infrequent [9].

Frequent Pattern Mining is the process of finding frequent patterns of activities from the input sensor data stream [3][5]. FP - growth mines the complete set of frequent itemsets without candidate generation which adopts a divide and conquers strategy.

First, it compresses the database into a fp-tree. A frequent pattern tree is a prefix-tree structure storing frequent patterns for the transaction database, where the support of each tree node is no less than a predefined minimum support threshold. [10][9].

High Utility Pattern Mining [6] is another important data mining technique that is being used in predicting the user behavior and alert the abnormal condition of the residents. HUP (High Utility Pattern) mining mines the patterns whose utility must be greater than the user specified threshold.

Interactive Mining is another important technique in which repeated mining with different minimum support thresholds can be performed by making use of the same data structure wherein a property such as "build once mine many" property is utilized.

These data mining techniques of smart home helps the inhabitants in centralizing the management [11] and services in a home more effectively and provide them all required functions in order to exchange internal information and enables to keep in touch with the outside world.

These techniques also helps the person in optimizing his / her living style, organize the day-to-day schedule, securing a high quality of living condition and in turn helps the person to reduce bills from a variety of energy consumptions in a house.

## 3. Proposed Work

Initially, the proposed work mainly focuses on modeling the behavior of the individuals and determine how to deal with them. Our approach of Activity Modeling consists of two phases 1) to extract the frequent patterns of activities using frequent pattern mining 2) modeling the frequent activity patterns identified. Fig.1 shows the Activity modeling which uses FPM (Frequent Pattern Mining) and HUP (High Utility Pattern) mining to detect the activities.

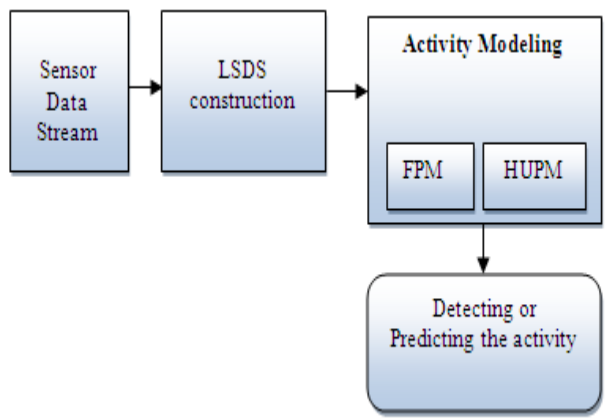

Fig. 1 Activity Modeling in Smart Home

Fig.1 shows the Activity Modeling being carried out at smart home. With the sensor data stream as input, LSDS (Linked Sensor data stream) is constructed using Algorithm 1. Then activity modeling is developed by using FPM ( Frequent pattern mining) and HUPM(High Utility Pattern mining). Then the intelligent system predicts or detects the activity. If it finds any anomaly, the system alerts the care giver.

A data stream is a collection of unbounded data that arrive in order of time. Frequent patterns are the activities that appear in a data stream with frequency no less than the user specified threshold.

| Sensor Id / Sensor data at Time $T_k$ | A | B | C | D | E |
|---|---|---|---|---|---|
| $T_1$ | 0 | 0 | 0 | 1 | 1 |
| $T_2$ | 1 | 0 | 1 | 0 | 0 |
| $T_3$ | 0 | 1 | 1 | 1 | 0 |
| $T_4$ | 1 | 1 | 0 | 0 | 0 |
| $T_5$ | 0 | 1 | 0 | 0 | 1 |
| $T_6$ | 1 | 1 | 0 | 1 | 0 |
| $T_7$ | 1 | 1 | 1 | 0 | 0 |
| $T_8$ | 1 | 1 | 0 | 1 | 1 |

Fig. 2  Sensor Data Stream.

For example, a set of activities in cooking such as {rinsing the rice, combining rice with water, switching on the stove, boiling the rice} that appear together in a sensor data stream is a frequent pattern. First sensor data stream collected in the smart home is analyzed to look for such frequent patterns that represents the activities. Fig.2 shows the example of events collected in a smart home which represents the status (ON / OFF) of the sensors using binary data at a certain time duration. In Fig.2 at time T1, the sensors D and E are triggered.

### 3.1 LSDS Construction

Here in Fig. 2 the Sensor data stream is represented in matrix format which increases the complexity in terms of both space and time and also needs repeated scan of the database to find out the frequent patterns. Even though a particular event is not occurred it is being represented as zeros which results in unused spaces.

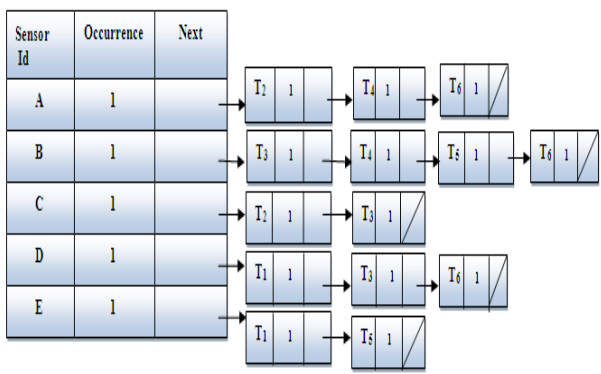

Fig. 3 Linked Sensor Data Stream($W_1$).

To overcome this issue, this sensor data stream can be represented by a new structure called LSDS (Linked Sensor Data Stream) as shown in Fig.3 which is the LSDS for Window$_1$. This new structure is constructed using Algorithm 1. And the set of sensor events are retrieved using Algorithm 2.

*Algorithm 1 Description: Construction of LSDS*

Algorithm.1 shows the procedure to construct the Linked Sensor Data Stream. The line 8 tests whether the current batch number exceeds the window size. Here it is assumed that the window size as 3. It means that only three batches of information can be inserted into the FPS - tree. It is assumed that each batch consists of two sensor data at time $T_k$. If the condition is true, the line 10 calls the Delete procedure. Otherwise, the Insert procedure is called.

*Algorithm 1 : Construction of LSDS*

1  **Procedure** LSDS Construction ($T_k$, $S_k$, j, M, N)
2  $T_k$ is the time duration between sensor data.
3  $S_k$ is the set of events at $T_k$.
4  j is the current batch number.
5  M is the number of batches in a window.
6  N is the number of sensor data in a window.
7  **begin**
8      **foreach** batch $B_j$ **do**
9          if j > M **then**
10             Call delete $B_j$
11             **foreach** Time $T_k$ in batch $B_j$ do
12                 Insert $B_j$ into FPS – tree
13             **end**
14         **end**
15         **else**
16             **foreach** Time $T_k$ in batch $B_j$ **do**
17                 **foreach** sensor event detected $S_k$ in $T_k$ **do**
18                     Call Insert $T_k$
19                     Insert $B_j$ into FPS – tree
20                 **end**
21             **end**
22         **end**
23     **end**
24  **end**

### 3.2 Definition

Let $S_n$ {$s_1,s_2,s_3.....s_n$} be the set of events collected form smart home and minimum support threshold as $M_s$. then $S_n$ is frequent if and only if the set of events $(S_n) >= M_s$.

*Algorithm 2 Description: Sensor Events at Time $T_k$*

Algorithm 2 describes the procedure to identity the sensor events at time $T_k$. Here, the element is the set of set of events at time $T_k$ which is to be retrieved. Initially, we set element as NULL. The if condition in lines 6 - 8 tests whether the received sensor data is empty or not. The procedure returns when $T_k$ is empty. For each event in the linked representation of Sensor data stream (LSDS), a set is created that consists of the set of transactions which is non – empty. If we perform intersection between the time duration $T_k$ and the set of sensor events which is non empty and if that condition is true, then the element is retrieved.

*Algorithm 2 : Identification of Sensor Events at Time $T_k$*

1  **Procedure** SensorEventsatTime $T_k$ (Tk,LSDS)
2  $T_k$ is the time duration between sensor data.

```
3   LSDS is the Linked Sensor Data Stream
4   begin
5       Set element = NULL
6       if T_k is NULL then
7           Return
8       end
9       else
10          foreach sensor data i in LSDS do
11              Create a set S_i = Set of sensor events of
                        "i" in LSDS
12              if(Tk ∩ Si) == true then
13                  element ← element ∪ sensor data i
14              end
15          end
16      end
17      Return element
18  end
```

*Algorithm 3 Description: Sensor Events occurring together*

Using Algorithm 3 it is possible to find out the set of activities that are frequent. This reduces the candidate generation problem. Here an intersection operation is being carried out to find the sensors of activities that occurs together. The main advantage of Algorithm 3 is that it reduces repeated scan on the database.

3.3 Architecture of FPM (Frequent Pattern Mining)

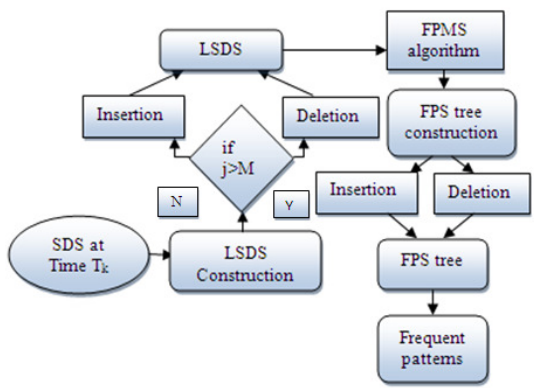

Fig. 4 Architecture of FPM

*Algorithm 3 : Identification of sensor events occurring together*

**1 Procedure** SensorEventsTogetheratTime
   $T_k(S_1…S_k, LSDS)$
**2** $S_1..S_k$ is the set of sensor events to be extracted
**3** LSDS is the Linked Sensor Data Stream
**4** j is the index

```
5   val is the set of sensors in which events that occurs
    together
6   begin
7       if (S_1…S_k) is NULL then
8           Return
9       end
10      else
11          foreach sensor data i in LSDS do
12              foreach j in (S_1…S_k) do
13                  if (S_1…..S_k) == i then
14                      Create a set S_i = Set of sensor
                            events of "i" in LSDS
15                  end
16              end
17          end
18      end
19      foreach sensor data i in LSDS do
20          Set y = 0
21          if S_i is not empty   /* S_i is a set of sensor
                                events from previous events */
22          then
23              if (y==0) then
24                  val = S_i
25                  Set y =1
26              end
27              else
28                  val = val ∩ S_i
29              end
30          end
31          if val!=empty then
32              (S_1….S_k) occurs together
33              Set of sensor events = {val}
34          end
35      end
36  end
```

Fig 4 shows the architecture of LSDS construction and FPS tree construction. Initially, sensor data stream as in Fig.2 is given as input for the construction of LSDS (Linked sensor data stream) using Algorithm 1. If the current batch number exceeds the window size, deletion is performed else insertion is done. After LSDS is construction, FPS tree is constructed using Algorithm 4 which results in frequent patterns of activities.

3.4 High Utility Pattern Mining

Fig 5 shows the process of applying utility to the sensors. As shown in Fig 5 each sensor is assigned a utility value or power consumed by them. Then the cumulative utility power is calculated. Then Frequent pattern tree which includes power (utility) consumption is constructed using Algorithm 4.

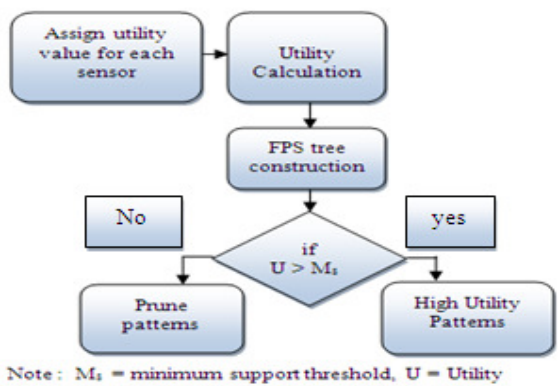

Fig. 5 High Utility Pattern Mining

If that final utility satisfies the threshold, they are taken as High utility patterns of activities. The patterns that do not satisfy the threshold are then pruned.

*Algorithm 4 Description: FPS tree Construction*

The insert procedure is described in lines . The delete procedure is described in lines . Here the sliding window approach [4][14] is used where the window contains some batches. After the first window, a new window is formed whenever a new batch of sensor events arrives.

The "if condition" in lines 5-7 tests whether the received sensor data is empty or not at time Tk. and it returns when Tk is empty. If the sensor - id of x matches with the sensor - id of any child node of R, then the twu value is updated. If there is no such match, the procedure creates a new child node and a counter array is being created. The insert procedure is recursively called with the remaining sensor data.

The delete procedure removes the batches of information by performing one time left shift operation. Then it updates the twu value in the header table. If the counter array contains all zero values, The delete procedure deletes node C from FPS - tree.s

*Algorithm 4 : FPS Tree Construction*

1  Procedure InsertFPStree($T_k$,j,M,R)
2  $T_k$ is the time duration between sensor data., j is the current batch number.
3  M is the number of batches in a window and R is the root of the current subtree.
4  **begin**
5     **if** $T_k$ is NULL **then**
6        | Return
7     **end**
8     Let divide $T_k$ as [x|X], where x is the first element and X is the remaining list.
9     if R has a child C such that C.sensor-id = x.sensor-id then
10       if j > M **then**
11          | C.twu[M] = C.twu[M] + x.twu
12       end
13       else
14          | C.twu[j] = C.twu[j] + x.twu
15       end
16    end
17    else
18       create a new node C as child of R
19       C.sensor-id = x.sensor-id
20       create a twu counter array with length M for C and initialize the array locations with zero
21       if j > M **then**
22          | C.twu[M] = x.twu
23       end
24       else
25          | C.twu[j] = x.twu
26       end
27    end
28    Call InsertFPStree(X,j,M,C)
29 **end**
30 Procedure DeleteFPStree(R)
31 R is the root of the current subtree
32 **begin**
33    **if** R is NULL **then**
34       | return
35    **end**
36    foreach child C of R do
37       Perform one time left shift operation in the twu counter of C
38       Update twu value in the header table H
39       Call DeleteFPStree(C)
40       if twu counter array of C contains all zero values then
41          delete C from FPS-tree
42       end
43    end
44 **end**

## 4. Experimental Results

The Experimental results as shown in Fig 6 shows the frequent patterns of Activities of Daily Living (ADLs) of the resident which is being detected by the smart environment and used in Activity Modeling..

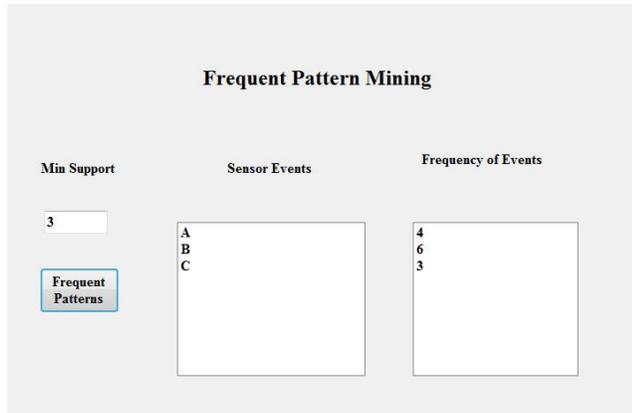

Fig. 6 Frequent Pattern Mining

Fig 7 shows the high utility patterns of Activities of Daily Living (ADLs) of the resident which utilizes extreme power (utility) at the home environment. Such information about anomaly detection is informed to the user or caregiver.

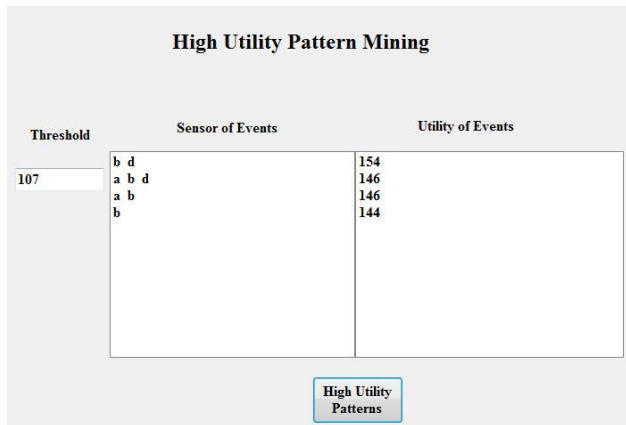

Fig. 7 High Utility Pattern Mining

## 5. Conclusion

Thus smart home used the data mining techniques such as Frequent pattern mining and High Utility Pattern mining and mined the activity patterns of the residents from the collected sensor data. The proposed algorithms increased the efficiency of frequent pattern mining. The extended work focused on detecting the high usage of power consumption of the residents at home environments and also helps them in using the resources more efficiently.

**Menaka Gandhi . J**  Received B.Tech degree in IT from MSEC (Meenakshi Sundararajan Engg College),chennai in 2010. Worked as a Lecturer in SVCT for 1 year. Currently Pursuing M.E CSE in Sri Venkateswara College of Engineering (Batch : 2011 – 2013). Participated in workshop on "Membrane Computing" at SVCE, 2011. Secured first class topper in sem 1 of M.E and received a merit scholarship of Rs.22,500 /-. Also secured first class topper in sem 3 of M.E. Published paper in one International Conference and was awarded "Interscience Scholastic Award" for the Best paper and presentation.

**Gayathri K.S**  Received B.E degree in CSE from Madras University in 2001 and M.E degree from Anna University, Chennai. She is doing her Ph.D in the area of Reasoning in Smart Environments. Currently working as a Associate Professor in the Dept. of CSE, Sri Venkateswara College of Engineering (SVCE) with a teaching experience of 12 years. Published research papers in one National Conference and two International Conferences. Organized two workshops on Artificial Intelligence, Also a working member of DON (Data sciences, Open systems and Next generation Research Lab) Lab at SVCE.
.